\RequirePackage{fix-cm}
\documentclass[a4paper,11pt]{article}

\usepackage{authblk}
\usepackage[utf8]{inputenc}
\usepackage[T1]{fontenc}
\usepackage{pslatex}
\usepackage[english]{babel}
\usepackage{fancyhdr}
\usepackage{hyperref}
\usepackage{amsmath,amssymb,amsfonts,amsthm}
\usepackage{mathtools}
\usepackage{graphicx}
\usepackage{pgfplots}
\usepackage{algorithm}
\usepackage{algorithmic}
\usepackage{microtype}
\usepackage{rotating}
\usepackage{todonotes}
\usepackage{tikz}
\usepackage{tikz-qtree}
\usepackage{xparse}
\usepackage{appendix}
\usepackage{url}
\usepackage{tabularx}
\usepackage{paralist}
\usepackage{booktabs}
\usepackage{multirow}
\usepackage{subcaption}
\usepackage{cancel}
\usetikzlibrary{calc,shapes.callouts,shapes.arrows}
\usetikzlibrary{automata}
\usetikzlibrary{positioning}
\usetikzlibrary{decorations.pathreplacing}
\usetikzlibrary{decorations.pathmorphing}
\usetikzlibrary{arrows}
\usetikzlibrary{shapes}

\newcommand{\N}{\mathbb{N}}
\newcommand{\Z}{\mathbb{Z}}
\newcommand{\F}{\mathbb{F}}

\providecommand{\keywords}[1]{\textbf{\textit{Keywords }} #1}

\begin{document}

\title{On Counts and Densities of Homogeneous Bent Functions: An Evolutionary Approach}

\author[1]{Claude Carlet}
\author[2]{Marko \DH urasevic}
\author[2]{Domagoj Jakobovic}
\author[3]{Luca Mariot}
\author[2]{Stjepan Picek}
\author[4]{Alexandr Polujan}

\affil[1 ]{{\normalsize University of Bergen, Bergen, Norway}

    {\small \texttt{claude.carlet@gmail.com}}}

\affil[2 ]{{\normalsize Faculty of Electrical Engineering and Computing, University of Zagreb, Unska 3, Zagreb, Croatia} \\

{\small \texttt{marko.durasevic@fer.hr, domagoj.jakobovic@fer.hr, stjepan@computer.org}}}

\affil[3 ]{{\normalsize Semantics, Cybersecurity and Services Group, University of Twente, 7522 NB Enschede, The Netherlands}

{\small \texttt{l.mariot@utwente.nl}}}

\affil[4 ]{{\normalsize Otto-von-Guericke University Magdeburg, Germany}
	
{\small \texttt{alexandr.polujan@ovgu.de}}}
	
\maketitle

\begin{abstract}
Boolean functions with strong cryptographic properties, such as high nonlinearity and algebraic degree, are important for the security of stream and block ciphers. These functions can be designed using algebraic constructions or metaheuristics. This paper examines the use of Evolutionary Algorithms (EAs) to evolve homogeneous bent Boolean functions, that is, functions whose algebraic normal form contains only monomials of the same degree and that are maximally nonlinear. We introduce the notion of density of homogeneous bent functions, facilitating the algorithmic design that results in finding quadratic and cubic bent functions in different numbers of variables.
\end{abstract}


\section{Introduction}
\label{sec:introduction}

Boolean functions with specific cryptographic features have been studied extensively within symmetric cryptography for many years~\cite{carlet_2021}. Designs for both stream ciphers and block ciphers typically require functions that are balanced, highly nonlinear, of a prescribed algebraic degree, or efficient to evaluate, among other desirable attributes. Such functions can be obtained with different approaches. One possibility is to rely on algebraic constructions, where entire families of Boolean functions are generated with guaranteed properties. However, deriving these constructions is often challenging, and for several classes of functions, no algebraic method is currently known, making empirical search a viable alternative.

Empirical strategies usually fall into three broad categories: random search, ad-hoc heuristics, and metaheuristics.  
Random sampling of the search space is straightforward but seldom competitive except in trivial instances.  
Specialised heuristics can perform remarkably well, sometimes producing state-of-the-art results~\cite{4167738}, but their success depends strongly on the problem structure and the availability of expert knowledge for their design.  
Metaheuristics provide a middle ground, offering good performance across a variety of problems while typically requiring only limited problem-specific tuning~\cite{Djurasevic2023}.

In this work, we consider a particularly demanding problem for which the number of known solutions is very limited. More precisely, our focus is on evolving \emph{homogeneous} Boolean functions, i.e., functions whose algebraic normal form (ANF) contains only monomials of the same degree. Such functions are interesting from an implementation perspective, as their uniform degree can simplify and speed up evaluation.

Only a small number of studies have addressed the construction of homogeneous Boolean functions so far. The earliest contribution, dating back to 2001, used combinatorial techniques to obtain balanced and bent homogeneous functions in six variables~\cite{10.1007/3-540-48970-3_3}.  
Quadratic homogeneous bent functions (of degree 2) are well understood and characterised in classical literature~\cite{MacWilliams-Sloane}.  
For the cubic case (of degree 3), however, there is essentially a single known construction~\cite{Seberry2000ConstructionOC}.  
In contrast to the homogeneous case, the broader problem of evolving bent functions has been examined in numerous works employing a range of methods, e.g.,~\cite{FullerDM03,10.1007/978-3-319-10762-2_41,10.1145/2908812.2908915,manzoni20,mariot22}.  
Related research has also demonstrated that evolutionary approaches can successfully produce Boolean functions with high algebraic degree and other cryptographically relevant criteria~\cite{00190,cryptoeprint:2013:011,10.1007/978-3-319-10762-2_81}.

In this paper, we investigate the efficiency of several Boolean function representations based on symbolic, truth table, and algebraic normal form encodings.
We experiment with Boolean function sizes from dimension 6 to dimension 12 and try to evolve homogeneous bent Boolean functions.
Previous results indicate the problem is relatively easy for degree 2, but very difficult for degree 3~\cite{CarletEvolving}.
Our contributions are twofold: first, we offer theoretical insights about the notion of the density of homogeneous bent functions, which predicts the likelihood of finding these functions depending on the number of terms in the ANF form.
Second, with a custom algorithm design and careful analysis, we can find cubic bent functions, a result that has, up to now, not been achieved with EAs.

\section{Preliminaries}
\label{sec:background}
We denote by $\F_2=\{0,1\}$ the finite field with two elements, equipped with XOR and logical AND, respectively, as the sum and multiplication operations. The $n$-dimensional vector space over $\F_2$ is denoted as $\F_2^n$, consisting of all $2^n$ binary vectors of length $n$. Given $a, b \in \F_2^n$, their inner product equals $a\cdot b = \bigoplus_{i=1}^{n} a_{i}b_{i}$ in $\mathbb F_{2}^n$. A Boolean function of $n$ variables is a mapping $f: \F_2^n \to \F_2$. Interested readers can find detailed information about Boolean functions in~\cite{MacWilliams-Sloane,carlet_2021}.

\subsection{Boolean Function Representations}

\paragraph{Truth Table Representation.}
The most basic means to uniquely represent a Boolean function $f: \F_2^n \to \F_2$ is by using its truth table. The truth table of a Boolean function $f$ is the list of pairs $(x, f(x))$ of input vectors $x \in \F_2^n$ and function outputs $f(x) \in \F_2$. Once a total order has been fixed on the input vectors of $\F_2^n$ (most commonly, the lexicographic order), the truth table can be identified only by the $2^n$-bit function output vector.

\paragraph{Walsh-Hadamard Transform.}
The Walsh-Hadamard transform $W_{f}: \F_2^n \to \Z$ is another commonly used unique representation of a Boolean function $f: \F_2^n \to \F_2$, which allows us to characterize several interesting cryptographic properties. Formally, the Walsh-Hadamard transform measures the correlation between $f$ and the linear functions $a\cdot x$, for all $a \in \N$, as follows:
\begin{equation}
W_{f} (a) = \sum\limits_{x \in \mathbb{F}_{2}^{n}} (-1)^{f(x) \oplus a\cdot x},
\end{equation}
with the sum calculated in ${\mathbb Z}$. The Walsh-Hadamard transform is an involution up to a normalization by a constant. Therefore, one can retrieve the truth table representation of $f$ from the spectrum of its Walsh-Hadamard coefficients $W_f(a)$.

\paragraph{Algebraic Normal Form.}
A third unique representation of a Boolean function $f: \F_2^n \to \F_2$ is as a multivariate polynomial in the quotient ring
$\mathbb{F}_{2}\left[x_{1},..., x_{n}\right]/(x_{1}^{2} \oplus x_{1},..., x_{n}^{2} \oplus x_{n})$. This polynomial is the Algebraic Normal Form (ANF) of $f$, and it is defined as:
\begin{equation}
f(x) = \bigoplus_{\substack{a \in \mathbb{F}_{2}^{n}}} h(a)\cdot x^{a},
\end{equation}  
\noindent
where $h(a)$ is given by the binary M\"{o}bius transform:
\begin{equation}
h(a)= \bigoplus_{\substack{x \preceq a}}  f(x), \text{ for any } a \in \mathbb{F}_2^n,
\end{equation}
with $\preceq$ denoting the covering relation between vectors of $\F_2^n$, i.e., $a$ covers $x$ means that $x_i \leq a_i, \forall i \in \left\lbrace 0, \ldots, n-1 \right\rbrace$.
Similarly to the Walsh-Hadamard transform, the M\"{o}bius transform is also an involution, so one can use it to switch between the truth table and the algebraic normal form representations of a function.

\subsection{Properties and Bounds}
\label{sec:boolean_properties}

\paragraph{Nonlinearity}
The minimum Hamming distance between a Boolean function $f$ and all affine functions is the nonlinearity of $f$, which is calculated from the Walsh-Hadamard spectrum as follows~\cite{carlet_2021}:
\begin{equation}
\label{eq:nonlinearity}
nl_{f} = 2^{n - 1} - \frac{1}{2}\max_{a \in \mathbb{F}_{2}^{n}} \left \{ |W_{f}(a)| \right \}.
\end{equation}
For every $n$-variable Boolean function, $f$ satisfies the covering radius bound:
\begin{equation}
\label{eq_boolean_covering}
    nl_{f} \leq 2^{n-1}-2^{\frac n 2 - 1}.
\end{equation}
Notice that Eq.~\eqref{eq_boolean_covering} cannot be tight when $n$ is odd. 

\paragraph{Bentness.}

The functions whose nonlinearity equals the maximal value from Eq.~\eqref{eq_boolean_covering} are called bent. Bent functions exist for $n$ even only.

\paragraph{Algebraic Degree.}
The algebraic degree $deg_f$ of a Boolean function \textit{f} is defined as the number of variables in the largest product term of the function's ANF having a non-zero coefficient, see~\cite{MacWilliams-Sloane}:
\begin{equation}
deg_f = \max \{w_H(a) : \ a \in \F_2^n, \ h(a) = 1 \}.
\end{equation} 

The algebraic degree of a bent function cannot exceed $n/2$.
A Boolean function is affine if and only if it has an algebraic degree at most 1. We will call quadratic functions the Boolean functions of algebraic degree at most 2, and cubic functions those of algebraic degree at most 3. This means that an affine function is a particular quadratic function (in the same sense that a constant function is a particular affine function).

\paragraph{Homogeneity.}
A Boolean function is called homogeneous if all the monomials in its algebraic normal form have the same algebraic degree. 
The only known homogeneous bent functions are quadratic and cubic ones. Furthermore, it is not known whether homogeneous bent functions of higher degrees exist~\cite{Polujan2020}. 

\section{Related Work}
\label{sec:related}

Research on the use of evolutionary algorithms and other metaheuristics to design Boolean functions with strong cryptographic properties dates back to the 90s. In this section, we give only a brief overview of the most significant work in this field, referring the reader to the survey~\cite{Djurasevic2023} for a more complete account.

Millan et al.~\cite{10.1007/BFb0028471} were the first to pioneer the use of Genetic Algorithms (GA) to evolve highly nonlinear Boolean functions, focusing on the truth table as a genotype representation for the candidate solutions, and performing experiments on functions with 8 to 16 inputs. Later works in the same direction explored more refined genetic operators---such as crossover operators that preserve the balancedness property~\cite{10.1007/BFb0054148,manzoni20})---or the optimization of other cryptographic properties beyond nonlinearity, such as autocorrelation and correlation immunity~\cite{10.1007/3-540-36231-2_20}. Concerning other representations, Clark et al.~\cite{1299941} proposed the spectral inversion method, in which the Walsh spectra corresponding to specific cryptographic profiles are optimized using a Simulated Annealing algorithm to obtain Boolean functions with optimal properties. Later, Mariot and Leporati~\cite{10.1007/978-3-319-26841-5_3} devised a GA to design semi-bent functions through the spectral inversion approach.

Besides GA and Simulated Annealing, other metaheuristics have been considered for the design of Boolean functions for cryptographic purposes. Picek et al.~\cite{10.1145/2464576.2464671} were the first to experiment with GP to evolve 8-variable Boolean functions with a good combination of high nonlinearity and other desirable cryptographic properties, observing that it substantially outmatched GA's performance. Later, Picek et al. also proposed to use CGP to evolve Boolean functions with good cryptographic properties~\cite{10.1007/978-3-319-16501-1_16}. Mariot and Leporati~\cite{10.1145/2739482.2764674} considered a discrete version of Particle Swarm Optimization (PSO) to evolve functions from 7 to 12 variables having a good trade-off of nonlinearity, correlation immunity, and propagation criteria. 

A related research strand focuses on the use of EA to evolve bent functions specifically. In this regard, Picek and Jakobovic applied GP to evolve algebraic constructions, which are, in turn, used to generate bent functions of up to 24 variables~\cite{10.1145/2908812.2908915}. 
Hrbacek and Dvorak~\cite{10.1007/978-3-319-10762-2_41} considered the use of Cartesian GP to synthesize bent functions up to 16 variables, while Husa and Dobai investigated linear GP for the same problem, obtaining bent functions up to 24 inputs~\cite{10.1145/3067695.3084220}. Picek et al. adapted various EAs, including GA and GP, to evolve quaternary bent functions, which are a generalization of classic bent functions over the alphabet $\mathbb{Z}_4$~\cite{picek18a}. Mariot et al. investigated the use of EA to design hyper-bent Boolean functions~\cite{hyperbent}, a subclass of binary bent functions that lie at maximum distance from all bijective monomial functions. Further, Mariot et al.~\cite{mariot22} applied evolutionary strategies to design algebraic constructions based on cellular automata to generate quadratic bent functions.

As mentioned in Section~\ref{sec:introduction}, there are very few works in the literature addressing the construction of homogeneous Boolean functions, such as~\cite{10.1007/3-540-48970-3_3}, which provides a combinatorial classification of bent and balanced homogeneous functions in 6 variables. As far as we know, the only work applying EA to the design of homogeneous bent functions is~\cite{CarletEvolving}, which provides preliminary results on the use of GA and GP to evolve this type of function. Interestingly, the authors managed to evolve quadratic homogeneous bent functions, but no examples of cubic homogeneous functions were found. This empirical finding serves as the starting point for the present work, in which we aim to improve on the results of~\cite{CarletEvolving}.

\section{Methodology}
\label{sec:methodology}

\subsection{Counts and Densities of  Homogeneous Bent Functions}
\label{sec:bent_count}

In this section, we present the exact counts and empirical densities of bent quadratic and cubic homogeneous Boolean functions, focusing on the two completely characterized cases $n = 6$ and $n = 8$. We then show that these results highlight the increasing difficulty of evolving cubic homogeneous bent functions compared to quadratic ones, thus addressing the question recently raised in~\cite{CarletEvolving}. We use these insights to inform our evolutionary approach in the next sections.

To this end, we introduce the notion of the \emph{density} of homogeneous bent functions of algebraic degree $d$ in $n$ variables, defined as  
\begin{equation}\label{eq:density}
	\delta_{n,d} = \frac{|\mathcal{HB}_{n,d}|}{2^{\binom{n}{d}}},
\end{equation}
where $|\mathcal{HB}_{n,d}|$ denotes the number of homogeneous bent functions of degree $d$ in $n$ variables, and $2^{\binom{n}{d}}$ is the total number of homogeneous Boolean functions of degree $d$.  
A more detailed density measure is obtained by restricting the count to functions containing exactly $k$ terms, namely  
\begin{equation}\label{eq:density k terms}
	\delta_{n,d,k} = \frac{|\mathcal{HB}_{n,d,k}|}{\binom{\binom{n}{d}}{k}},
\end{equation}
where $|\mathcal{HB}_{n,d,k}|$ denotes the number of homogeneous bent functions of degree $d$ in $n$ variables with $k$ terms, and $\binom{\binom{n}{d}}{k}$ is the number of all homogeneous Boolean functions of degree $d$ with $k$ terms.  

In the following two subsections, we compare these density values for quadratic ($d = 2$) and cubic ($d = 3$) homogeneous bent functions in $n = 6$ and $n = 8$ variables, since only for these cases are all such functions completely known.

\subsubsection{Quadratic Homogeneous Bent Functions} For quadratic homogeneous bent functions in $n$ variables, the value $\delta_{n,2}$ defined by Eq.~\eqref{eq:density} can be computed theoretically, since the value $|\mathcal{HB}_{n,2}|$ is well-known (see, e.g.,~\cite[Chapter 15]{MacWilliams-Sloane}) and is given by:
\begin{equation}\label{eq:bent-count}
	\bigl|\mathcal{HB}_{n,2}\bigr| = 2^{k^2-k}\prod_{i=0}^{k-1}\bigl(2^{2i+1}-1\bigr).
\end{equation}

Now, we are interested both in the asymptotic value of $\delta_{n,2}$ and in the exact behavior for small $n$ (here $n=6,8$) of the values $\delta_{n,2,k}$.  To compute  $\displaystyle \lim_{n\to\infty}\delta_{n,2}$, one can use Wolfram Mathematica~\cite{Mathematica}, which gives
\begin{equation}
	\lim_{n\to\infty}\delta_{n,2} = \left(\frac12; \frac14\right)_\infty \approx 0.419422,
\end{equation}
where $(a; q)_{\infty}$ denotes the $q$-Pochhammer symbol. This result means that roughly $42\%$ of all quadratic homogeneous functions in $n$ variables are bent when $n\to\infty$. 

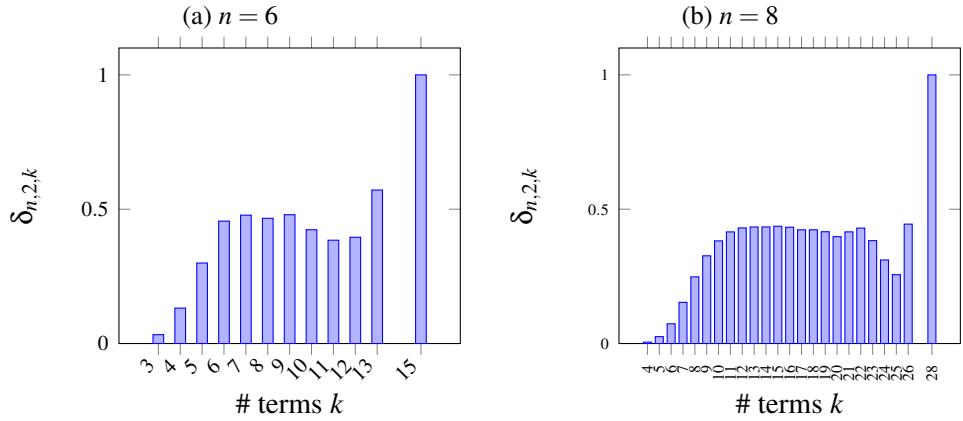
\begin{figure}[ht]
\centering
\begin{subfigure}{0.48\textwidth}
\centering
\caption{$n=6$}
\begin{tikzpicture}
\begin{axis}[
    ybar,
    width=\textwidth,
    height=5.5cm,
    bar width=4pt,
    enlarge x limits=0.15,
    xlabel={\# terms $k$},
    ylabel={$\delta_{n,2,k}$},
    xtick=data,
    ymin=0,
    xticklabel style={rotate=45, anchor=east},
    ticklabel style={font=\scriptsize}
]
\addplot+ coordinates {
    (3,0.032967) (4,0.131868) (5,0.299700) (6,0.455544)
    (7,0.477855) (8,0.466200) (9,0.479520) (10,0.423576)
    (11,0.384615) (12,0.395604) (13,0.571429) (15,1)
};
\end{axis}
\end{tikzpicture}
\end{subfigure}\hfill
\begin{subfigure}{0.48\textwidth}
\centering
\caption{$n=8$}
\begin{tikzpicture}
\begin{axis}[
    ybar,
    width=\textwidth,
    height=5.5cm,
    bar width=3pt,
    enlarge x limits=0.1,
    xlabel={\# terms $k$},
    ylabel={$\delta_{n,2,k}$},
    xtick=data,
    ymin=0,
    xticklabel style={rotate=90, anchor=east},
    ticklabel style={font=\tiny}
]
\addplot+ coordinates {
    (4,0.00512821) (5,0.025641) (6,0.0735786) (7,0.153238)
    (8,0.248167) (9,0.326543) (10,0.382071) (11,0.415514)
    (12,0.430312) (13,0.433775) (14,0.433969) (15,0.436428)
    (16,0.432903) (17,0.42329) (18,0.423306) (19,0.416094)
    (20,0.397791) (21,0.415655) (22,0.429729) (23,0.382906)
    (24,0.311111) (25,0.25641) (26,0.444444) (28,1.0)
};
\end{axis}
\end{tikzpicture}
\end{subfigure}
\caption{Non-zero density values $\delta_{n,2,k}$.}
\label{fig: quadratic homogeneous}
\end{figure}

For $n = 6, 8$, all quadratic homogeneous bent functions can be enumerated within minutes. Using this data, we compute the values of $\delta_{n,2,k}$ and present them as histograms in Figure~\ref{fig: quadratic homogeneous}. For both $n = 6$ and $n = 8$, we observe that for most values of $k$, the densities $\delta_{n,2,k}$ are close to the asymptotic value $\delta_{n,2}$ given by Eq.~\eqref{eq:density}. From an evolutionary perspective, this indicates that a large part of the search space can be effectively explored: functions with a broad range of term counts can be evolved without a substantial drop in success probability, as bent functions are relatively evenly distributed across most values of $k$.

\subsubsection{Cubic Homogeneous Bent Functions}
Unlike the quadratic case, the number of homogeneous cubic bent functions in $n$ variables is not known theoretically. Therefore, the asymptotic value of $\delta_{n,3}$ cannot be determined. However, for $n = 6$ and $n = 8$, the exact counts of homogeneous cubic bent functions are known, which allows us to compute the corresponding values of $\delta_{n,3}$. Specifically, it was shown in~\cite{CRB} that the total number of homogeneous cubic bent functions in six variables is $|\mathcal{HB}_{6,3}| = 30$. For eight variables, the total number is $|\mathcal{HB}_{8,3}| = 293{,}760$; see~\cite{MengClassifyingHomogeneous} for details. Using these values, it is straightforward to compute:
\begin{equation}
		\delta_{6,3} = \frac{30}{2^{\binom{6}{3}}}\approx 2.86102\times 10^{-5}\quad\mbox{and}\quad
		\delta_{8,3} = \frac{293{,}760}{2^{\binom{8}{3}}}\approx 4.07674\times 10^{-12}.
\end{equation}

These results show that, unlike in the quadratic case, cubic homogeneous bent functions are extremely rare among all cubic homogeneous functions. Moreover, they are not evenly distributed across functions with a fixed number of terms $k$. For instance, in the case of $n = 6$ variables, all $30$ homogeneous cubic bent functions contain exactly $16$ terms in their ANF~\cite{CRB}, hence $\delta_{6,3,k} = 0$ for all $k \neq 16$. Similarly, for $n = 8$, only a few values of $k$ satisfy $\delta_{8,3,k} \neq 0$; these are listed in Table~\ref{tab:cubic-n8}.

\begin{table}[ht]
	\centering
	\caption{Counts and densities for cubic homogeneous bent functions in $n=8$.}
	\label{tab:cubic-n8}
	\begin{tabular}{c|c|c}
		\toprule
		\# terms $k$ & $ \; |\mathcal{HB}_{8,3,k}| \; $ & $\delta_{n,3,k}$ \\
		\midrule
		24 & 6{,}720   & $1.54304\times 10^{-12}$ \\
		27 & 13{,}440  & $1.81992\times 10^{-12}$ \\
		28 & 5{,}760   & $7.53070\times 10^{-13}$ \\
		32 & 6{,}720   & $1.54304\times 10^{-12}$ \\
		34 & 13{,}440  & $6.27280\times 10^{-12}$ \\
		35 & 19{,}200  & $1.42564\times 10^{-11}$ \\
		36 & 80{,}640  & $1.02646\times 10^{-10}$ \\
		37 & 67{,}200  & $1.58246\times 10^{-10}$ \\
		39 & 40{,}320  & $4.11439\times 10^{-10}$ \\
		41 & 40{,}320  & $2.48073\times 10^{-9{\phantom0}}$ \\
		\bottomrule
	\end{tabular}
\end{table}

For larger values of $n$, a complete enumeration of cubic homogeneous bent functions remains infeasible. Nevertheless, the known values of $k$ for which such functions exist can be used. These values, derived from previous studies~\cite{CRB,Polujan2020}, are summarized in Table~\ref{tab:kvalues}. 

\begin{table}[ht]
	\caption{Known values of $k$ s.t. homogeneous cubic bent functions with $k$ terms in the ANF in $n$ variables exist.}
	\label{tab:kvalues}
	\centering
	\begin{tabular}{c|l}
		\hline
		$n$ & Values of $k$ \\
		\hline
		10 & $39, 49, 53, 57, 58, 61, 65, 66, 69, 70, 72, 75, 78$ \\
		12 & $60, 90, 100, 110, 130, 140, 150$ \\
		16 & $168$ \\
		\hline
	\end{tabular}
\end{table}

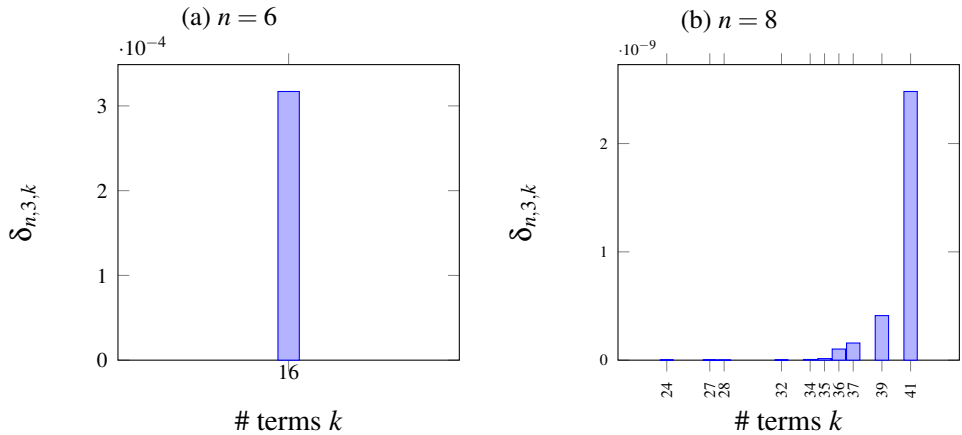
\begin{figure}[ht]
\centering

\begin{subfigure}{0.48\textwidth}
\centering
\caption{$n=6$}
\begin{tikzpicture}
\begin{axis}[
    ybar,
    width=\textwidth,
    height=5.5cm,
    bar width=8pt,
    enlarge x limits=0.3,
    xlabel={\# terms $k$},
    ylabel={$\delta_{n,3,k}$},
    xtick=data,
    ymin=0,
    xticklabel style={rotate=0, anchor=center},
    ticklabel style={font=\scriptsize},
    yticklabel style={/pgf/number format/fixed}
]
\addplot+ coordinates { (16,3.17e-4) };
\end{axis}
\end{tikzpicture}
\end{subfigure}\hfill
\begin{subfigure}{0.48\textwidth}
\centering
\caption{$n=8$}
\begin{tikzpicture}
\begin{axis}[
    ybar,
    width=\textwidth,
    height=5.5cm,
    bar width=5pt,
    enlarge x limits=0.2,
    xlabel={\# terms $k$},
    ylabel={$\delta_{n,3,k}$},
    xtick=data,
    ymin=0,
    xticklabel style={rotate=90, anchor=east},
    ticklabel style={font=\tiny},
    yticklabel style={/pgf/number format/fixed}
]
\addplot+ coordinates {
    (24,1.54304e-12) (27,1.81992e-12) (28,7.5307e-13) (32,1.54304e-12)
    (34,6.2728e-12) (35,1.42564e-11) (36,1.02646e-10) (37,1.58246e-10)
    (39,4.11439e-10) (41,2.48073e-9)
};
\end{axis}
\end{tikzpicture}
\end{subfigure}
\caption{Non-zero density values $\delta_{n,3,k}$.}
\label{fig: cubic homogeneous}

\end{figure}

To highlight the difference between quadratic and cubic homogeneous bent functions, as well as their distributions among homogeneous functions with a fixed number of terms $k$ in the ANF, we present in Figure~\ref{fig: cubic homogeneous} a histogram summarizing the results from Table~\ref{tab:cubic-n8}. 

In contrast to Figure~\ref{fig: quadratic homogeneous}, the viable $k$-region for cubic homogeneous bent functions is very narrow, indicating that most of the cubic search space contains almost no bent functions. From an evolutionary perspective, this extreme sparsity, combined with the enormous size of the cubic search space, explains the practical difficulty of evolving cubic homogeneous bent functions.

\subsection{Encodings}

\subsubsection{Symbolic Encoding (GP).}
The first encoding in our experiments uses tree-based genetic programming (GP) to represent a Boolean function in its symbolic form. 
This encoding usually achieves the best results when dealing with the evolution of Boolean functions with cryptographic properties~\cite{Djurasevic2023}.
In this case, we represent a candidate solution with a tree whose leaves correspond to the input variables $x_1,\ldots, x_n \in \F_2$. The internal nodes are Boolean operators that combine the inputs received from their children and forward their output to the respective parent nodes. 
We use the following function set: OR, XOR, AND, AND2, XNOR, IF, and the NOT function that takes a single argument. This function set is common in previous applications and is based on our tuning results.
The output of the root node is the output value of the Boolean function. The truth table of the function $f: \F_2^n \to \F_2$ is determined by evaluating the tree over all possible $2^n$ assignments of the inputs at the leaves. 

To restrict the search to homogeneous functions of a given degree, we perform the following additional transformations: the truth table is first converted to ANF via the M\"obius transform, then all the elements of the ANF corresponding to monomials of the "wrong" degrees are set to zero. 
The obtained \textit{corrected ANF} is transformed back into a truth table and finally into the Walsh-Hadamard spectrum, based on which we calculate nonlinearity.

\subsubsection{Truth Table Encoding (TT).}
The most common option for encoding a Boolean function is the truth table (TT) encoding~\cite{Djurasevic2023}, represented with a bitstring. 
For a Boolean function with $n$ inputs, the truth table is encoded as a bitstring of length $2^n$.
The bitstring represents the Boolean function upon which the algorithm operates directly. Therefore, the algorithm, in this case, explores the full space of $n$-variable Boolean functions, which has size $2^{2^n}$. 
As in the previous encoding, in each evaluation, the truth table is transformed into the ANF form with the M\"obius transform, and monomials not corresponding to a desired degree are reset to force homogeneity.
The corrected ANF is converted back to a truth table, after which the nonlinearity properties are evaluated.

\subsubsection{Reduced ANF Encoding (rANF).}
The truth table encoding introduced above does not guarantee that the function the evolutionary algorithm is working with is always homogeneous. Hence, we also considered an encoding based on the ANF, where an individual is represented as a bitstring of length $\binom{n}{d}$, with $d$ being the target degree. Each component in this bitstring specifies whether the corresponding $d$-degree monomial occurs (1) in the ANF of the function or not (0). Standard genetic operators can be applied to this bitstring, and the individual is then converted to the full ANF representation by mapping only to the bits of the appropriate $d$-degree monomials. In this way, the evolutionary algorithm only explores the restricted search space of homogeneous Boolean functions, which has size $2^{\binom{n}{d}}$. Each individual is evaluated by first retrieving its truth table and then by computing its nonlinearity through the Walsh-Hadamard transform.

\subsubsection{Weighted ANF Encoding (wANF).}
Finally, based on the densities of cubic (of degree 3) bent homogeneous functions from Section~\ref{sec:bent_count}, we adopt the approach in which we use the reduced ANF encoding, but also restrict the number of monomials to a specific value.
All the previous encodings may be viewed as "unrestricted", since the number of monomials may vary. 
On the other hand, in \textit{weighted ANF}, encoding this number is preselected to one of the values for which bent functions of degree 3 are shown to exist.
With this encoding, we perform a separate experiment for each of the predicted values of the number of monomials for a given function size.
Rather than penalizing solutions with "wrong" number of monomials, the evolutionary genotype is designed so that the designated number of ones in the bitstring remains the same during evolution, with customized genetic operators that preserve this property.

\subsection{Fitness Functions}

\subsubsection{(Homogeneous) Bent Functions.}

Several objective functions can be defined to find bent functions, regardless of the representation and search algorithm. 
The approach used here is based on common choices in related works~\cite{Djurasevic2023} and the authors' previous experience.
Apart from maximizing the nonlinearity value, the applied objective value considers the whole Walsh-Hadamard spectrum and not only its extreme value (see Eq.~\eqref{eq:nonlinearity}); hence, we count the number of occurrences of the maximal absolute value in the spectrum, denoted as $\#max\_values$.
As higher nonlinearity corresponds to a lower maximal absolute value, we aim for as few occurrences of the maximal value as possible, hoping it gets easier for the algorithm to reach the next nonlinearity value.
The algorithm is thus provided with additional information, making the objective space more gradual. The fitness function to optimize for bent functions is defined as:
\begin{equation}
\label{eq:bent}
fit_{bent} = nl_{f} + \frac{2^n - \#max\_values}{2^n}.
\end{equation}
The second term never reaches the value of $1$ since, in that case, we effectively reach the next nonlinearity level.
Since all encodings are either constrained or repaired to represent a homogeneous function of a given degree, the homogeneity property is not explicitly included in the fitness function.

\subsubsection{Homogeneous Bent Functions with $k$ Monomials.}

The previous fitness function optimizes for bent functions with any number of monomials in the ANF form; since it may be beneficial to restrict the number of monomials in the cubic case, we employ the additional function that penalizes the distance to the desired monomial number $k$. Only if the number of monomials is equal to $k$, the nonlinearity property is evaluated and awarded:
\begin{equation}
\label{eq:bentk}
fit_{bent, k} = \left\{ \begin{aligned} & 
    -\lvert num\_monomials - k \rvert \text{, if } num\_monomials \neq k; \\ & 
    nl_{f} + \frac{2^n - \#max\_values}{2^n} \text{, otherwise.} \\ \end{aligned} \right.
\end{equation}
Obviously, this fitness function is applicable only to the first three encodings, since the weighted ANF (wANF) already has the preselected number of monomials.

\subsection{Algorithms and  Parameters}
\label{sec:settings}

\paragraph{Common Experimental Parameters.}
We employ the same evolutionary algorithm for all encodings: a steady-state selection with a 3-tournament elimination operator (denoted SST). 
In each iteration of the algorithm, three individuals are chosen at random from the population for the tournament, and the worst one in terms of fitness value is eliminated. 
The two remaining individuals in the tournament are used with the crossover operator to generate a new child individual, which then undergoes mutation with individual mutation probability $p_{mut} = 0.5$. The mutated child replaces the eliminated individual in the population.
All experiments use a population size of 500 individuals and the same stopping criterion of $10^6$ evaluations.

\paragraph{Genetic Operators.}
Concerning the genetic operators for both the bitstring and restricted encodings, we use the simple bit mutation and the shuffle mutation. For crossover, we employ one-point and uniform crossover operators. Each time the evolutionary algorithm invokes a crossover or mutation operation, one of the previously described operators is randomly selected.

In the case of weighted ANF encoding, the individual bitstrings are always initialized with $k$ bits set to one. 
The crossover and mutation operators used in this work are based on~\cite{manzoni20}, so that all genetic operators preserve the number of ones throughout the evolution. 
For the crossover, we used the balanced weighted crossover operator, and the mutation used a two-bit inversion (which flips two random bits) and a mixing mutation, which shuffles the genes between two randomly chosen positions in the bitstring.

For the symbolic encoding, the genetic operators used in our experiments with tree-based GP are simple tree crossover, uniform crossover, size fair, one-point, and context preserving crossover~\cite{poli08:fieldguide} (selected at random), and subtree mutation.
The option to use multiple genetic operators was based on previous results indicating better convergence when using a diverse set of operators.

\paragraph{Local Search.}
Finally, all three encodings can also be used with a generic local search operator, which works in the following way: the operator acts on a single solution and performs a predefined number of mutations. 
If a better solution is found, the new solution immediately replaces the original one, and the operator is applied again.
If no better solution is found after the given number of mutations, the operator terminates.
The operator is applied after each generation and acts upon the current best solution and a number of random solutions.
In our experiments, the number of solutions undergoing local search was set to 1\% of the population size, and the number of trials (random mutations per individual) was set to 30.

\section{Experimental Results}
\label{sec:results}

In the experiments, all configurations were executed in 30 runs, and relevant statistical values are reported.
For the quadratic case (degree 2), we used only the "unrestricted" encodings (GP, TT, rANF), which do not limit the number of monomials in the ANF form. 
This problem turns out to be relatively simple: all encodings managed to find a bent homogeneous function in all problem sizes (number of variables from 6 to 12), and in all runs.
The results for this case are therefore not elaborated any further.

A much more difficult (and interesting) problem is the evolution of cubic bent functions (of degree 3). 
Previously used evolutionary approaches~\cite{CarletEvolving} were unable to find even a single cubic bent function in the sizes we consider.
In six variables, all the encodings managed to find cubic bent functions, with most encodings succeeding in every run; the overall results are concisely presented as success rates over 30 runs for each configuration and are shown in Table~\ref{tab:bent3}.

\begin{table}[h!]
\centering
\small
\begin{tabular}{c c | c c c c | c c}
\hline
\textbf{vars} & \textbf{weight} 
& \textbf{ GP } & \textbf{ TT } & \textbf{rANF} & \textbf{wANF}
& \textbf{rANF/LS} & \textbf{wANF/LS} \\
\hline
\multirow{2}{*}{6} 
& unrestricted  & 21 & 30 & 30 & -- &  &  \\ 
& 16  & 24 & 30 & 30 & 30 &  &  \\ 
\hline
\multirow{11}{*}{8}
& unrestricted  & 0 & 0 & 0 & -- & 0 & -- \\
& 24  & 0 & 0 & 0 & 0 & 0 & 0 \\
& 27  & 0 & 0 & 0 & 0 & 0 & 0 \\
& 28  & 0 & 0 & 0 & 0 & 0 & 0 \\
& 32  & 0 & 0 & 0 & 0 & 0 & 0 \\
& 34  & 0 & 0 & 0 & 0 & 0 & 0 \\
& 35  & 0 & 0 & 0 & 0 & 0 & 0 \\
& 36  & 0 & 0 & 0 & 0 & 0 & 0 \\
& 37  & 0 & 0 & 0 & 1 & 0 & 1 \\
& 39  & 0 & 0 & 1 & 4 & 0 & 2 \\
& 41  & 0 & 0 & 1 & 4 & 0 & 2 \\
\hline
\end{tabular}
\caption{Number of successful runs (out of 30) for each encoding and configuration.}
\label{tab:bent3}
\end{table}

The column "restricted weight" denotes whether the first fitness function $fit_{bent}$ was used, which corresponds to the entry "unrestricted". 
If the second fitness function $fit_{bent,k}$(\ref{eq:bentk}) was used, the entry in the column shows the preselected weight (the number of monomials), based on analysis in section~\ref{sec:bent_count}.
For cubic functions of 6 variables, this number can only be 16.
Indeed, even in the unrestricted experiments (with $fit_{bent}$), all the solutions contained exactly 16 monomials, which was shown by analyzing the functions after the evolution.

The GP encoding exhibited the worst performance, which is not in accordance with previous results in the literature when evolving cryptographically relevant Boolean functions.
However, the main reason for this is most likely the indirect relationship between the genotype (symbolic function form) and the resulting truth table; since in the decoding process all extraneous monomials (not corresponding to the selected degree) are removed, the function that is evaluated for nonlinearity is not identical to the one used internally in GP.

\begin{figure}[ht]
\centering
    \begin{subfigure}[b]{0.49\textwidth}
        \centering
    \includegraphics[trim=4cm 4cm 0cm 0cm, width=1\linewidth]{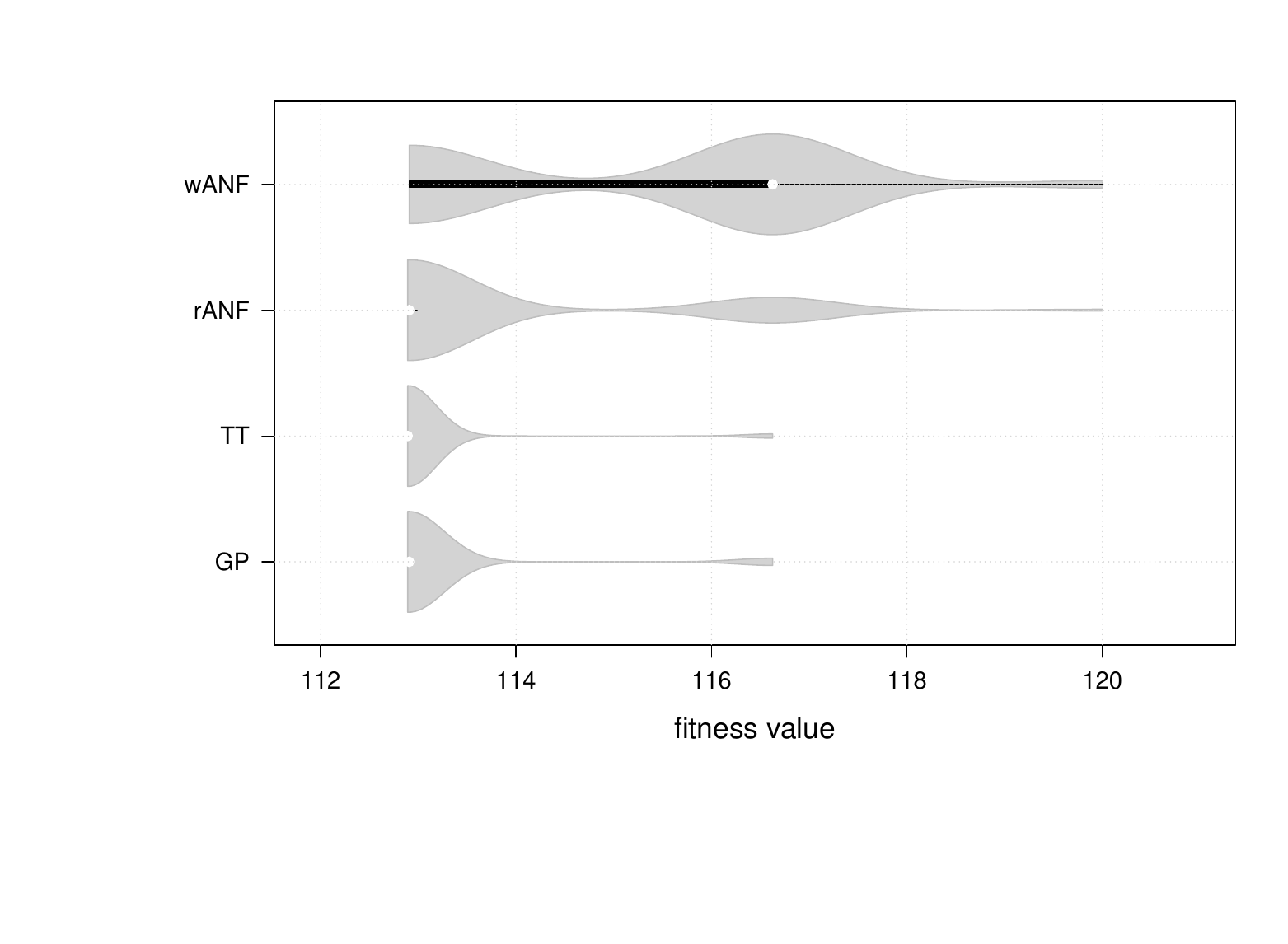}
    \caption{8 variables}
    \label{fig:size8}
    \end{subfigure}
    \hfill
    \begin{subfigure}[b]{0.49\textwidth}
        \centering
    \includegraphics[trim=4cm 4cm 0cm 0cm, width=1\linewidth]{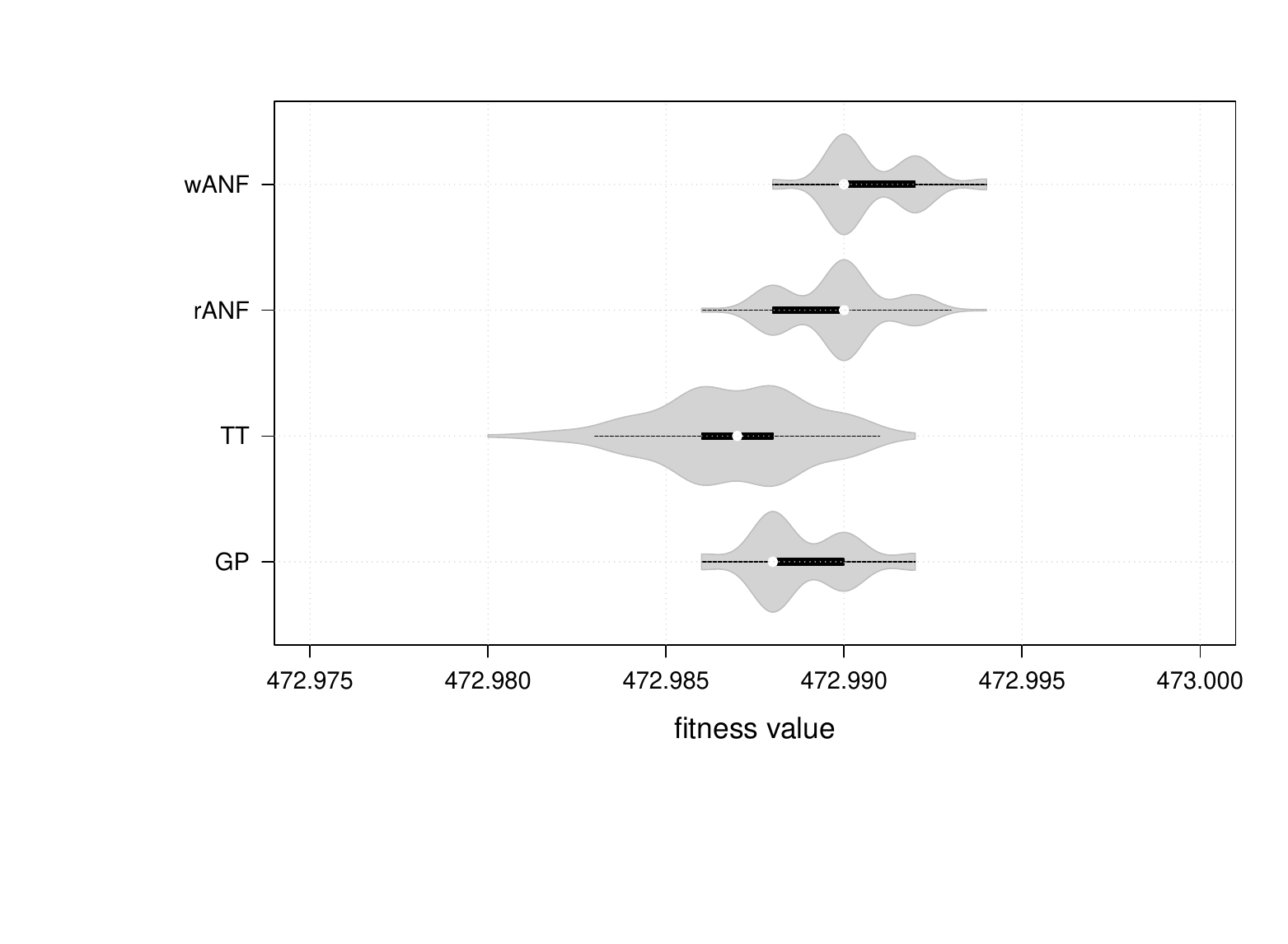}
    \caption{10 variables}
    \label{fig:size10}
    \end{subfigure}

    \vskip\baselineskip
    \begin{subfigure}[b]{0.49\textwidth}
        \centering
    \includegraphics[trim=4cm 4cm 0cm 0cm, width=1\linewidth]{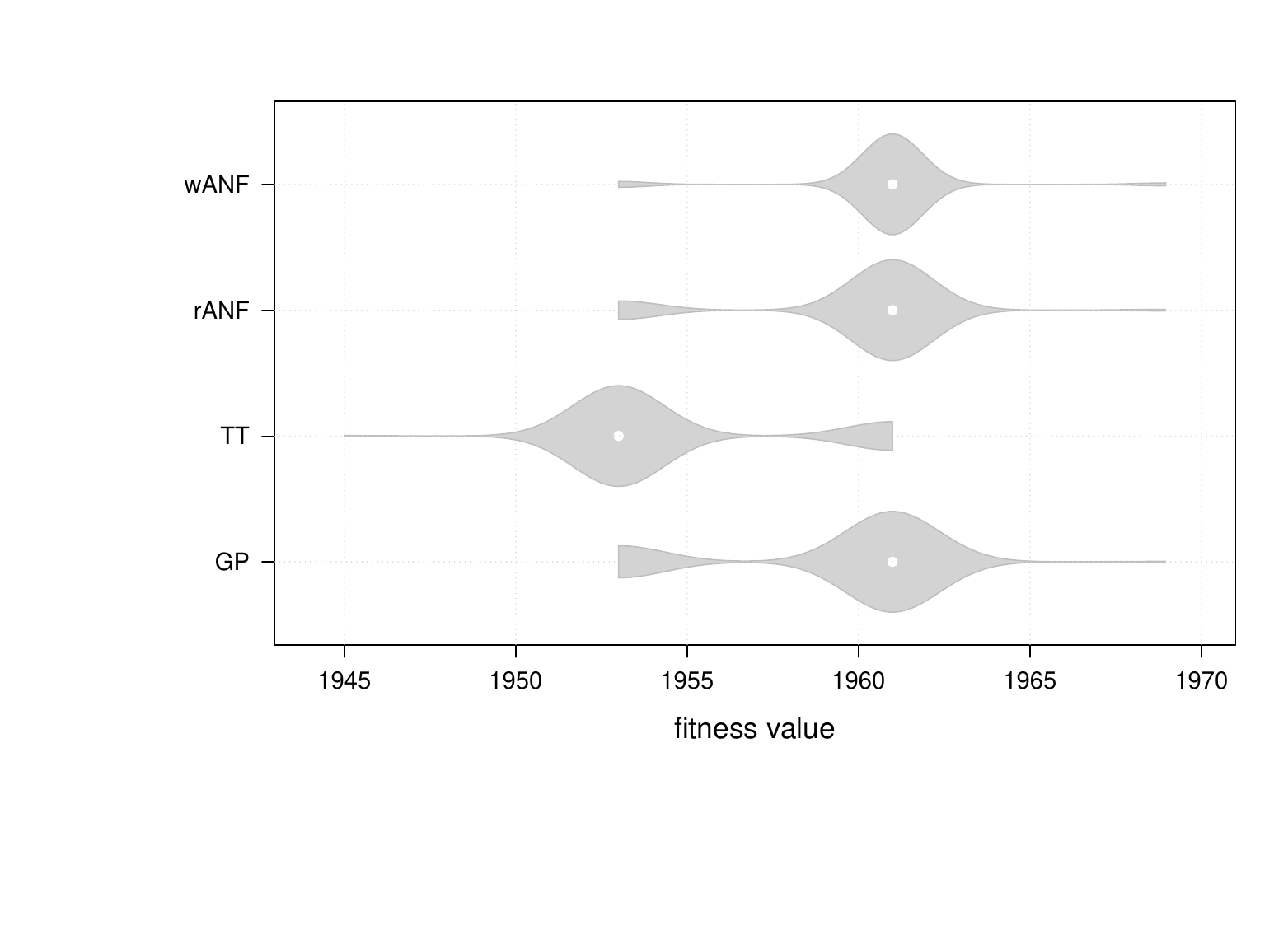}
    \caption{12 variables}
    \label{fig:size12}
    \end{subfigure}

\caption{Accumulated fitness values for all encodings, sizes 8-12.}
\end{figure}

For larger sizes, the algorithms only managed to find cubic bent functions in 8 variables (success rates for 10 and 12 are zero); the number of hits over 30 runs is presented in the lower part of Table~\ref{tab:bent3}.
Only the reduced ANF and weighted ANF encodings managed to find bent functions in a small number of runs. 
Notably, the successful runs were only achieved when bitstring weight was restricted, either by penalization in rANF or by design in wANF.
Apart from the number of successful runs, we may also observe absolute fitness values of different encodings over 30 runs; the accumulated best-of-run fitness values (over unrestricted and all restricted configurations) for $n=8$ are shown in Figure~\ref{fig:size8}.

We may investigate encoding efficiency by examining fitness values directly. The results were analyzed using the Kruskal-Wallis test to determine the differences and a Dunn post-hoc test with Bonferroni correction and significance level of 0.05. Furthermore, pairwise comparisons were conducted with Mann-Whitney U tests.
The results of the statistical tests show significant differences among all the encodings; subsequent pairwise comparisons suggest that TT is the worst performer, followed by GP, then by reduced ANF, and finally the weighted ANF as the best one.

The results for sizes 10 and 12 are presented in Figures~\ref{fig:size10} and~\ref{fig:size12}.
In 10 variables, the tests show statistically significant differences in the same order of efficiency, with TT being the worst and wANF the best method.
However, these differences are inconsequential, since all fitness values are grouped very tightly (note the scale in Figure~\ref{fig:size10}); disregarding the differences in fitness value, all the obtained results correspond to Boolean functions with the same nonlinearity value (which is the integer part of the fitness).
Finally, for 12 variables, the statistical differences are again present, but they are of little importance since no encoding is even close to the nonlinearity of bent functions in 12 variables (2016).

\subsubsection{Experiments with Local Search.}
To try to improve results, we included the local search operator for the best two encodings (rANF and wANF) and repeated the experiments for 8, 10, and 12 variables in the hope of finding cubic bent functions.
The statistical tests show that there are no significant differences between the local search and the original variants for both encodings; this observation also holds for all problem sizes.
Unfortunately, the number of successful runs with local search is even lower, which is evident from Table~\ref{tab:cubic-n8}, and there were no hits for larger sizes.

\subsubsection{Effect of Restricting the Number of Monomials.}
As mentioned before, the only configurations in which cubic bent functions in 8 variables are found are those that restrict the bitstring weight $k$.
In accordance with findings in Section~\ref{sec:bent_count}, bent functions are indeed found for weights that exhibit the largest densities of those functions (see Figure~\ref{fig: cubic homogeneous}).
In other words, it is very likely that without this restriction, those results would not be obtained.
The statistical tests also confirm significant differences in pairwise comparison between the unrestricted and all accumulated restricted configurations, showing the advantage of restricting the number of monomials.

\section{Conclusions}
\label{sec:conclusions}

The results show that evolving homogeneous quadratic bent functions is relatively straightforward: all encodings reached optimal solutions across all tested sizes. In contrast, the cubic case was substantially more challenging. While all encodings succeeded in finding cubic bent functions in six variables, only the reduced ANF and weighted ANF found solutions for eight variables, and none succeeded for larger sizes. Statistical analysis confirmed  performance differences between encodings, with the weighted ANF consistently outperforming all others, with the TT and GP encodings showing the weakest results. The inclusion of a local search operator did not improve success rates and, in some cases, reduced them. An important observation is that successful runs in the cubic case occurred only when the number of monomials was explicitly restricted, consistent with the theoretical distribution of bent functions over different weights. These results suggest that both the choice of encoding and the enforcement of structural constraints are crucial for guiding EAs towards feasible regions of the search space.

\bibliographystyle{abbrv}
\bibliography{bibliography}

\end{document}